\documentclass[10pt,twocolumn,letterpaper]{article}

\usepackage{cvpr}
\usepackage{times}
\usepackage{epsfig}
\usepackage{graphicx}
\usepackage{amsmath}
\usepackage{amssymb}
\usepackage{subfigure}
\usepackage[ruled,vlined]{algorithm2e}

\usepackage[breaklinks=true,bookmarks=false]{hyperref}

\cvprfinalcopy % *** Uncomment this line for the final submission

\def\cvprPaperID{****} % *** Enter the CVPR Paper ID here

\begin{document}

%%%%%%%%% TITLE
\title{Learning Robust Deep Face Representation}

\author{Xiang Wu\\
University of Science and Technology Beijing\\
Beijing, China\\
{\tt\small alfredxiangwu@gmail.com}}

\maketitle
%\thispagestyle{empty}

%%%%%%%%% ABSTRACT
\begin{abstract}
With the development of convolution neural network, more and more researchers focus their attention on the advantage of CNN for face recognition task. In this paper, we propose a deep convolution network for learning a robust face representation. The deep convolution net is constructed by 4 convolution layers, 4 max pooling layers and 2 fully connected layers, which totally contains about 4M parameters. The Max-Feature-Map activation function is used instead of ReLU because the ReLU might lead to the loss of information due to the sparsity while the Max-Feature-Map can get the compact and discriminative feature vectors. The model is trained on CASIA-WebFace dataset and evaluated on LFW dataset. The result on LFW achieves \textbf{97.77\%} on unsupervised setting for \textbf{single net}.
\end{abstract}

\section{Introduction}
In the past years, with the development of convolution neural network, numerous vision tasks benefit from a compact representation learning via deep model from image data. The performance in various computer vision applications, such as image classification\cite{he2015delving}, object detection\cite{szegedy2014scalable}, face recognition\cite{sun2014deep, taigman2014deepface, yi2014learning} and so on, achieved great progress.

For the face verification task, the accuracy on LFW, a hard benchmark dataset, has been improved from $97\%$\cite{taigman2014deepface} to $99\%$\cite{sun2014deep} in recent year via deep learning model. The main frameworks for face verification are based on multi-class classification\cite{sun2014deepface, taigman2014deepface} to extract face feature vectors and then the vectors are further processed by classifiers or patch model ensembles. However, the probability models such as Joint Bayesian\cite{chen2012bayesian} and Gaussian Processing\cite{lu2014surpassing} are based on strong assumptions which may not make effect on various situations. Other methods\cite{hu2014discriminative, schroff2015facenet} are proposed to optimize verification loss directly for matching pairs and non-matching pairs. The disadvantage of these verification based methods is that it is difficult to select training dataset for negative pairs and the threshold in verification loss function is set manually. Moreover, the joint identification and verification constraint is used for optimizing the deep face model in \cite{sun2014deep, yi2014learning} and it is also difficult to set the trade-off parameter between identification and verification loss for multi-task optimization.

In this paper, we propose a deep robust face representation learning framework. We utilize convolution networks and propose a Max-Feature-Map activation function, which the model is trained on CASIA-WebFace dataset\footnote{http://www.cbsr.ia.ac.cn/english/CASIA-WebFace-Database.html} and evaluated on LFW dataset.

The contributions of this paper are summarized as follows:
\begin{enumerate}
\item[(1)] We propose a Max-Feature-Map activation function whose values are not sparse while the gradients are sparse instead. The activation function can also be treated as the sparse connection to learn a robust representation for deep model.
\item[(2)] We build a \textbf{shallower single} convolution network and get better performance than DeepFace\cite{taigman2014deepface}, DeepID2\cite{sun2014deep} and WebFace\cite{yi2014learning}.
\end{enumerate}

The paper is organized as following. Section 2 briefly describes our convolution network framework and Max-Feature-Map activation function. We present our experimental results in Section 3 and conclude in Section 4.
\section{Architecture}

In this section, we describe the framework of our deep face representation model and the compact Max-Feature-Map activation function.
\subsection{Compact Activation Function}

\begin{figure}
\centering
\includegraphics[height=5.5cm, width=0.5\textwidth]{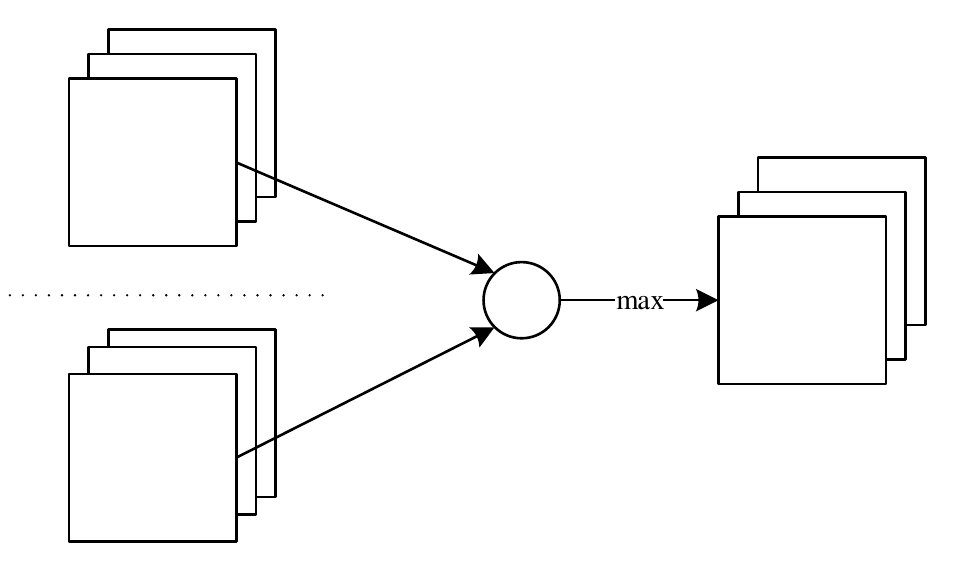}
\caption{Operation performed by Max-Feature-Map activation function}
\label{fig:activation}
\end{figure}

Sigmoid or Tanh is a nonlinear activation for neural network and often leads to robust optimization during DNN training\cite{hinton2006reducing}. But it may suffer from vanishing gradient when lower layers have gradients of nearly 0 because higher layer units are nearly saturate at -1 or 1. The vanishing gradient may lead to converge slow or poor local optima.

To overcome vanishing gradient, the Rectified linear unit(ReLU)\cite{nair2010rectified} offers a sparse representation. However, ReLU is at a potential disadvantage during optimization because the value is 0 if the unit is not active. It might lead to loss of some information especially for the first several convolution layers because these layers are similar to Gabor filter which both positive and negative responses are respected. To alleviate this problem, PReLU is proposed and it makes good effect on ImageNet classification task\cite{he2015delving}.

In order to make the representation compact instead of sparsity in ReLU, we propose the Max-Feature-Map(MFM) activation function which is inspired by \cite{goodfellow2013maxout}. Given an input convolution layer $C \in \mathbb{R}^{h\times w \times 2n}$, as is shown in Fig.\ref{fig:activation}, the Max-Feature-Map activation function can be written as
\begin{equation}
f=C^{k^{'}}_{ij}=\max_{1\leq k\leq n}(C^{k}_{ij}, C^{k+n}_{ij})
\end{equation}
where the number of convolution feature map $C$ is $2n$. The gradient of this activation function can be shown as
\begin{equation}
\frac{\partial{f}}{\partial{C^{k}}} = \left\{
    \begin{array}{l}
    1, \text{ if } C^{k}\geq C^{k+n}\\
    0, \text{ otherwise}
    \end{array}
\right.
\end{equation}

The Max-Feature-Map activation function is not a normal single-input-single-output function such as sigmoid or ReLU, while it is the maximum between two convolution feature map candidate nodes. This activation function can not only select competitive nodes for convolution candidates, but also make the $50\%$ gradients of convolution layers are 0. Moreover, the Max-Feature-Map activation function can also treated as the sparse connection between two convolution layers, which can encode the information sparsely onto a feature space.

\subsection{Convolution Network Framework}
\begin{figure*}
\centering
\includegraphics[height=4.5cm,width=\textwidth]{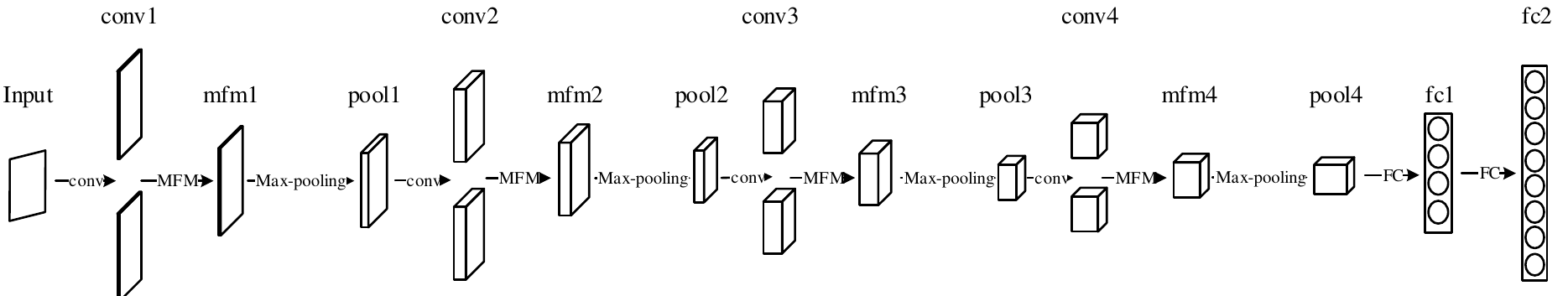}
\caption{An illustration of the architecture of our deep face convolution networks model.}
\label{fig:network}
\end{figure*}

\begin{table}
\centering
\caption{The architecture of the proposed deep face convolution network.}
\begin{tabular}{|c|c|c|c|}
\hline
Name & Type & \begin{tabular}{c}Filter Size\\/Stride\end{tabular} & Output Size \\
\hline
input & - & - & $144\times 144 \times 1$\\
\hline
crop & - & -& $128\times 128 \times 1$ \\
\hline
conv1\_1 & convolution & $9\times 9/1$ & $120\times 120 \times 48$ \\
conv1\_2 & convolution & $9\times 9/1$ & $120\times 120 \times 48$ \\
mfm1 & MFM & - & $120\times 120 \times 48$ \\
\hline
pool1 & max pooling & $2\times 2/2$& $60\times 60 \times 48$\\
\hline
conv2\_1 & convolution & $5\times 5/1$ & $56\times 56 \times 96$ \\
conv2\_2 & convolution & $5\times 5/1$ & $56\times 56 \times 96$ \\
mfm2 & MFM & - & $56\times 56 \times 96$ \\
\hline
pool2 & max pooling & $2\times 2/2$& $28\times 28 \times 96$\\
\hline
conv3\_1 & convolution & $5\times 5/1$ & $24\times 24 \times 128$ \\
conv3\_2 & convolution & $5\times 5/1$ & $24\times 24 \times 128$ \\
mfm3 & MFM & - & $24\times 24 \times 128$ \\
\hline
pool3 & max pooling & $2\times 2/2$& $12\times 12 \times 128$\\
\hline
conv4\_1 & convolution & $4\times 4/1$ & $9\times 9 \times 192$ \\
conv4\_2 & convolution & $4\times 4/1$ & $9\times 9 \times 192$ \\
mfm4 & MFM & - & $9\times 9 \times 192$ \\
\hline
pool4 & max pooling & $2\times 2/2$& $5\times 5 \times 192$\\
\hline
fc1 & fully connected & - & 256 \\
\hline
fc2 & fully connected & - & 10575 \\
\hline
loss & softmax & -& 10575 \\
\hline
\end{tabular}
\label{tab:network}
\end{table}

The deep face convolution network is constructed by four convolution layers, 4 max pooling layers, Max-Feature-Map activation functions and 2 fully connected layers as is shown in Fig.\ref{fig:network}.

The input image is $144\times 144$ gray-scale face image from CASIA-WebFace dataset. The detail parameters setting is presented in Table.\ref{tab:network}. We crop each input image randomly into $128\times 128$ patch as the input of the first convolution layer. The network include 4 convolution layers that each convolution layer is combined with two independent convolution parts calculated from the input. The Max-Feature-Map activation function and max pooling layer are used later. The fc1 layer is a 256-dimensional face representation since we usually consider that the face images usually lie on a low dimensional manifold and it is effective to reduce the complexity of the convolution neural network. The fc2 layer is used as the input of the softmax cost function and is set to the number of WebFace identities(10575). Besides, the proposed network has 4153K parameters which is smaller than DeepFace and WebFace net.

\section{Experiments}

\subsection{Data Pre-processing}

\begin{figure}
\centering
\subfigure[]{\includegraphics[height=3.5cm,width=0.2\textwidth]{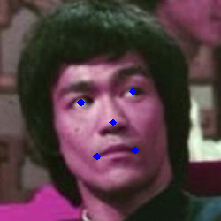}\label{fig:normalization(a)}}
\subfigure[]{\includegraphics[height=3.5cm,width=0.2\textwidth]{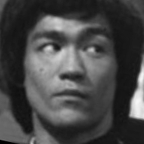}\label{fig:normalization(b)}}
\caption{Face image alignment for WebFace dataset. (a) is the facial points detection results and (b) is the normalization face image.}
\label{fig:normal}
\end{figure}

CASIA-WebFace dataset is used to train our deep face convolution network. It contains 493456 face images of 10575 identities and all the face images are converted to gray-scale and normalized to $144\times 144$ according to landmarks as is shown in Fig.\ref{fig:normalization(a)}. According to the 5 facial points extracted by \cite{sun2013deep} and manually revised, the distance between the midpoint of eyes and the midpoint of mouth is relative invariant to pose variations in yaw angle, therefore, it is fixed to 50 pixels and we also rotate two eye points horizontally to pos variations in roll angle. The normalization face image is shown in Fig.\ref{fig:normalization(b)}.

\subsection{Training Methodology}

To train the convolution network, we randomly select one face image from each identity as the validation set and the other images as the train set. The open source deep learning framework \textit{Caffe}\cite{jia2014caffe} is used for training the model.

The input for convolution network is the $144\times 144$ gray-scale face image and we crop the input image into $128\times 128$ and mirror it. These data augmentation method can improve the generalization of the convolution neural network and overcome the overfitting\cite{krizhevsky2012imagenet}.
Dropout is also used for fully connected layer and the ratio is set to 0.7.

Moreover, the weight decay is set to 5e-4 for convolution layer and fully connected layer except the fc2 layer. It is obvious that the fc1 face representation is only used for face verification tasks which is not similar to the image classification and objection task. However, the parameters of fc2 layer is very large. Therefore, it might lead to overfitting for learning the large fully-connected layer parameters. To overcome it, we set the weight decay of fc2 layer to 5e-3.

The learning rate is set to 1e-3 initially and reduce to 5e-5 gradually. The parameters initialization for convolution is Xavier and Gaussian is used for fully-connected layers. Moreover, the deep model is trained on GTX980 and the iteration is set to 2 million.
\begin{figure*}
\centering
\includegraphics[height=7.5cm,width=0.7\textwidth]{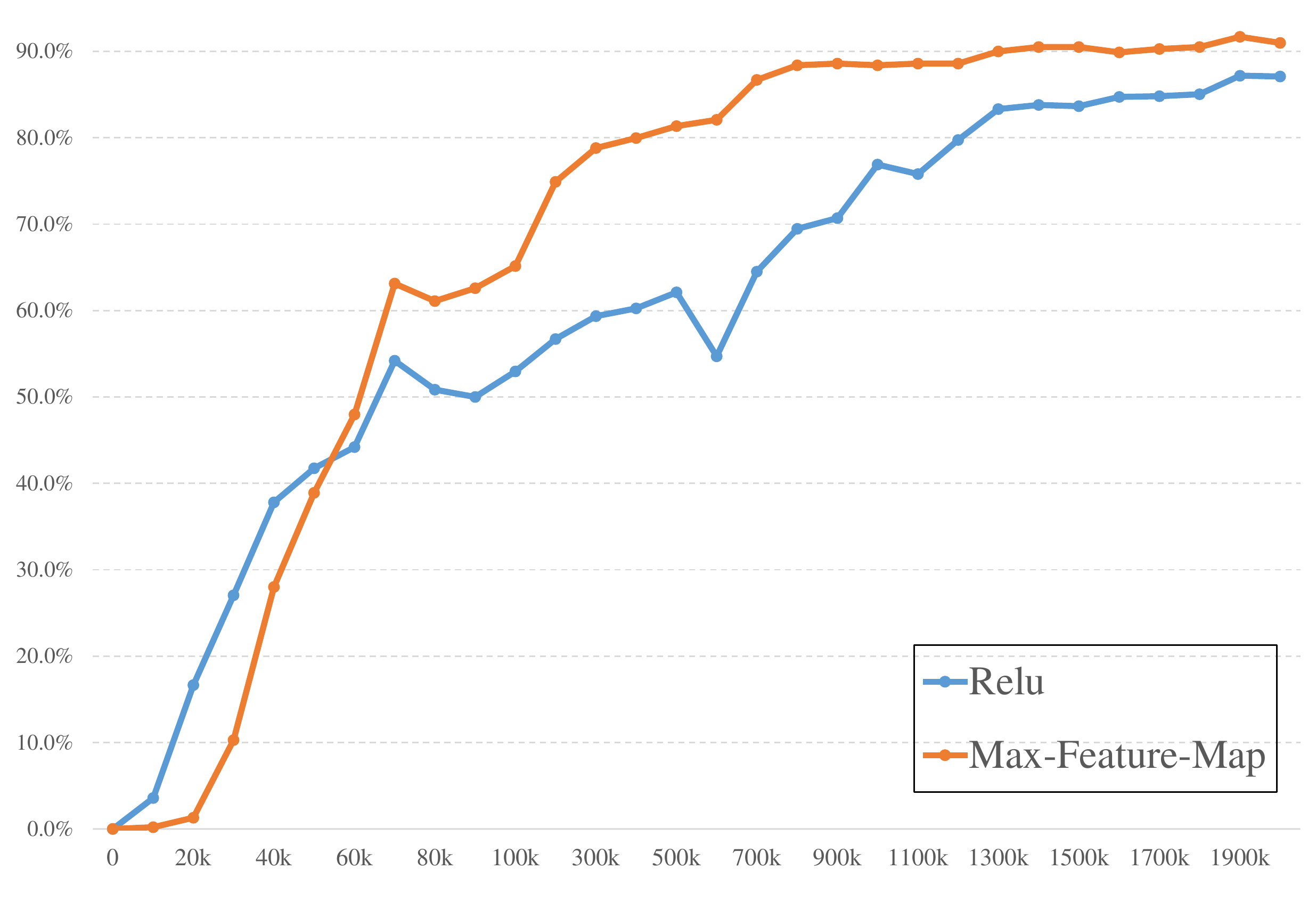}
\caption{Comparison with ReLU activation function and Max-Feature-Map activation function validation accuracy for CNN training.}
\label{fig:converage}
\end{figure*}

\subsection{Results on LFW benchmark}
The evaluation is performed on LFW dataset\footnote{http://vis-www.cs.umass.edu/lfw/} in detail. LFW dataset contains 13233 images of 5749 people for face verification. And all the images in LFW dataset are processed by the same pipeline as the training dataset and normalized to $128 \times 128$.

For evaluation, the face data is divided in 10 folds which contain different identities and 600 face pairs.
There are two evaluation setting about LFW training and testing: restricted and unrestricted.
In restricted setting, the pre-define image pairs are fixed by author (each fold contains 5400 pairs for training and 600 pairs for testing).
And in unrestricted setting, the identities of people within each fold for training is allowed to be much larger.

According to Fig.\ref{fig:converage}, compared with ReLU and Max-Feature-Map, the speed of convergence for Max-Feature-Map network is slower than ReLU due to the complexity of the activation and the randomness of initial parameters. However, with the progress of training, the validation accuracy for Max-Feature-Map net outperforms ReLU.

\begin{table}
\centering
\caption{The performance of our deep face model and compared methods on LFW.}
\begin{tabular}{|c|c|c|c|}
\hline
Method & \#Net & Accuracy & Protocol\\
\hline
DeepFace & 1 & 95.92\% & unsupervised\\
DeepFace & 1 & 97.00\% & restricted \\
DeepFace & 7 & 97.35\% & unrestricted\\
\hline
DeepID2 & 1 & 95.43\% & unsupervised \\
DeepID2 & 2 & 97.28\% & unsupervised\\
DeepID2 & 4 & 97.75\% & unsupervised\\
DeepID2 & 25 & 98.97\% & unsupervised\\
\hline
WebFace & 1 & 96.13\% & unsupervised \\
WebFace+PCA & 1 & 96.30\% & unsupervised \\
WebFace+Joint Bayes & 1 & 97.30\% & unsupervised \\
WebFace+Joint Bayes & 1 & 97.73\% & unrestricted \\
\hline
Our model(ReLU) & 1 & 97.45\% &unsupervised\\
Our model(MFM) & 1 & \textbf{97.77\%} & unsupervised\\
\hline
\end{tabular}
\label{tab:result}
\end{table}

We test our deep model performance via cosine similarity and ROC curve. The results\footnote{The model and configuration are released on my \href{https://github.com/AlfredXiangWu/face_verification_experiment}{Github}} are shown in Table.\ref{tab:result} and the EER on LFW achieves 97.77\%, which outperforms DeepFace\cite{taigman2014deepface}, DeepID2\cite{sun2014deep} and WebFace\cite{yi2014learning} for unsupervised setting\footnote{The unsupervised setting the model is not trained on LFW in supervised way.} for \textbf{single} net.
\section{Conclusions}

In this paper, we proposed a deep convolution network for learning a robust face representation. We use Max-Feature-Map activation function to learn a compact low-dimensional face representation and the results on LFW is 97.77\%, which the performance is the state-of-the-art on unsupervised setting for single net as far as we know.

{\small
\bibliographystyle{ieee}
\bibliography{sigproc}
}

\end{document}